\documentclass[11pt]{article}
\usepackage{coling2020}
\usepackage{times}
\usepackage{url}
\usepackage{latexsym}

\colingfinalcopy 

\usepackage{graphicx}
\usepackage{amsmath}
\usepackage{subcaption}
\usepackage{amsfonts}
\usepackage{tabulary,booktabs}
\usepackage{multirow}
\newcolumntype{L}[1]{>{\raggedright\let\newline\\\arraybackslash\hspace{0pt}}m{#1}}
\newcolumntype{C}[1]{>{\centering\let\newline\\\arraybackslash\hspace{0pt}}m{#1}}
\newcolumntype{R}[1]{>{\raggedleft\let\newline\\\arraybackslash\hspace{0pt}}m{#1}}

\title{Explain by Evidence: An Explainable Memory-based Neural Network for Question Answering}

\newcommand*{\affaddr}[1]{#1} 
\newcommand*{\affmark}[1][*]{\textsuperscript{#1}}
\newcommand*{\email}[1]{\texttt{#1}}

\author{%
\textbf{Quan Tran\affmark[1], Nhan Dam\affmark[2], Tuan Lai\affmark[3], Franck Dernoncourt\affmark[1]} \\ 
\textbf{Trung Le\affmark[2], Nham Le\affmark[4] and Dinh Phung\affmark[2]}\\
\affaddr{\affmark[1]Adobe Research  }
\affaddr{\affmark[2]Monash University}\\
\affaddr{\affmark[3] University of Illinois at Urbana-Champaign }
\affaddr{\affmark[4] University of Waterloo }\\
\email{\small \{qtran, franck.dernoncourt\}@adobe.com} \\
\email{\small \{nhan.dam, trunglm, dinh.phung\}@monash.edu} \\
\email{\small  tuanml2@illinois.edu, nhamlevan@gmail.com} \\
}

\date{}

\begin{document}
\setlength{\abovedisplayskip}{6pt}
\setlength{\belowdisplayskip}{6pt}
\maketitle
\begin{abstract}
  Interpretability and explainability of deep neural networks are challenging due to their scale, complexity, and the agreeable notions on which the explaining process rests. Previous work, in particular, has focused on representing internal components of neural networks through human-friendly visuals and concepts. On the other hand, in real life, when making a decision, human tends to rely on similar situations and/or associations in the past. Hence arguably, a promising approach to make the model transparent is to design it in a way such that the model explicitly connects the current sample with the seen ones, and bases its decision on these samples.
  Grounded on that principle, we propose in this paper an explainable, evidence-based memory network architecture, which learns to summarize the dataset and extract supporting evidences to make its decision. Our model achieves state-of-the-art performance on two popular question answering datasets (i.e. TrecQA and WikiQA). Via further analysis, we show that this model can reliably trace the errors it has made in the validation step to the training instances that might have caused these errors. We believe that this error-tracing capability provides significant benefit in improving dataset quality in many applications.
  
\end{abstract}

\section{Introduction}
\label{sec:intro}
\blfootnote{
    %
    %
    %
    %
    %
    %
    \hspace{-0.65cm}  
    This work is licensed under a Creative Commons 
    Attribution 4.0 International License.
    License details:
    \url{http://creativecommons.org/licenses/by/4.0/}.
}





Interpretability of neural networks is an active research field in 
machine learning. Deep neural networks might have tens if not hundreds of
millions of parameters~\cite{Devlin2018BERTPO,liu2019roberta} organized into 
intricate architectures. The sheer amount of parameters and the 
complexity of the architectures largely prevent human to directly make 
sense of 
which concepts and how the network truly learns. The comparative lack
of explainable intuition behind deep neural networks might hamper the 
development and adoption of those models. In certain scenarios, prediction accuracy alone
is not sufficient~\cite{caruana2015intelligible,lapuschkin2019unmasking}. For example, as 
discussed
in~\cite{zhang2018examining,zhang2018interpretable}, it is difficult
to trust a deep model even if it has high test set 
performance given the inherent biases in the dataset. Thus, we argue 
that interpretability is perhaps one of the keys to accelerate both
the development and adoption of deep neural networks.

There have been many 
successful attempts from the research community to make sense of deep 
models' prediction. These attempts can be broadly categorized 
into
several classes. One of the major classes concerns with 
the network visualization techniques, for example, visual saliency 
representations in convolutional models~\cite{simonyan2013deep,sundararajan2017axiomatic}. For recurrent neural networks (RNN), Karpathy et 
al.~\shortcite{karpathy2015visualizing} focused on
analyzing and visualizing the RNN to explain its 
ability to keep track of long-range information.

The visualization-based methods, although achieving great successes, 
still operate on a very high level of abstractions. It requires a 
great 
deal of machine learning knowledge to make use of those 
visualizations. Thus, these techniques are not always useful for
a broader audience, who might not have the machine learning expertise. Looking back at classic machine learning models, 
one class of models stands out as being very intuitive and
easy to understand: the instance-based learning algorithms. The $k$-nearest neighbors algorithm, a prime 
example, operates on a very 
human-like assumption. To elaborate, if the current circumstances are similar to that
of a known situation in the past, we may very well make this decision
based on the outcome of the past decision. We argue that this assumption
puts the interpretability on a much lower level of abstraction 
compared to the visualization methods. If somehow our model can learn 
how to link the evidences from the training data to the prediction 
phase, we will have a direct source of interpretability that can be
appreciated by a broader audience. 

The $k$-nearest neighbors algorithm, as an instance-based method, might not be a deep 
neural network technique; however, there have been many papers in
the deep model literature inspired by or related to this method. 
A notable example is the neural nearest neighbors 
network~\cite{plotz2018neural}. Moreover, there is a class of 
problems with strong links to $k$-nearest neighbors: few-shot learning.
It is from the two major papers in the few-shot learning
literature, the prototypical network~\cite{snell2017prototypical}
and the matching network~\cite{vinyals2016matching}, we find a 
potential realization for our ideas. 


In few-shot learning, it is possible to learn the support from each of the instances from the support set to the current prediction; however, such approach is infeasible when the training data get larger. Inspired by the techniques discussed in~\cite{ravi2016optimization}, we apply a training-level data summarizer based on the neural Turing machine (NTM)~\cite{graves2014neural} that reads the dataset and 
summarizes (or writes) it into a few meta-evidence nodes. These meta-evidence nodes, in turn,
lend support to each of the prediction similar to a few-shot learning 
model. The parameters of the NTM are jointly trained with other parameters of the network. Our final model not only has great
predictive power and achieves state-of-the-art results on two popular answer selection datasets, but also shows a strong ``error-tracing''
capability, in which the errors in the validation set can be traced
to the sources in the training set.

To summarize, our contributions in this work are twofold. First, we 
propose a novel neural network model that achieves state-of-the-art performance on two answer selection datasets. Second, we show the utility of the error-tracing
capability of our model to find the noisy instances in the training 
data that degrades the performance on the validation data. This capability
might be very useful in real-life machine learning scenarios where the
training labels are noisy or the inter-annotator agreement is low.

\section{Proposed Framework}
\label{sec:framework}

Question answering (or answer selection) is the task of identifying the correct answer to a question from a pool of candidate answers. It is an active research problem with applications in many areas \cite{tay2018hyperbolic,tayyar-madabushi-etal-2018-integrating,lai-etal-2018-review,rao-etal-2019-bridging,lai2020isa}. Similar to most recent papers on this topic ~\cite{Tay2018MultiCastAN,lai2019gated,garg2019tanda}, we cast the question answering problem as a binary classification problem by concatenating the question with each of the  
candidate answers and assigning positive label to the concatenation containing the correct answer.

In most supervised learning scenarios, performing a full distance calculation between the current data point and every training data point would be computationally intractable. To overcome this burden,
we propose a \textbf{memory controller} based on
NTM to summarize the dataset into \textbf{meta-evidence nodes}.  
Similar to NTM, the controller is characterized by \textbf{reading} and 
\textbf{writing mechanisms}. Assume that we provide the controller with $K$ cells $e_1,\ldots,e_K$ in a 
memory bank (i.e. to store $K$ support/evidence vectors), and let us 
denote the $t$-th data point as $x^t$ (obtained by 
using a pretrained embedding model to embed the concatenation of a 
question and a candidate answer), the memory controller works as 
follows.


\paragraph{Writing mechanism.} The writing mechanism characterizes
how the controller updates its memory given a new data point. To 
update the memory, however, we first
need an indexing mechanism for writing. Instead of using the original
indexing of the NTM, we adopt the simpler indexing procedure from 
the memory network, which has been proven to be useful in this task~\cite{lai2019gated}.
At time step $t$, for each incoming data point $x^{t}$, we compute the attention weight $w_{e_i^{t}}$ for the support vector $e_i^{t}$:
\begin{equation}
\label{eq:framework_att-weight-evidence}
    w_{e_i^{t}} = \frac{\exp\left(x^{t} \cdot e_i^t\right)}{\sum_{k=1}^K\exp\left(x^{t} \cdot e_k^t\right)}.
\end{equation}

From these attention weights, we find the writing index for an input 
$x^{t}$ by maximizing the cosine similarity between $x^{t}$ and the evidence vectors:
\begin{equation}
    k_{\rm best}^t = \arg \max_i w_{e_i^{t}}.
\end{equation}

With the writing index found, we compute the memory update weight via
a gating mechanism:
\begin{equation}
    g^t = \epsilon \times \sigma \left(\left(W_g \cdot e_{k_{\rm best}}^t + b_g\right) + \left(W_g \cdot x^t + b_g\right)\right),
\end{equation}
where $\epsilon$ is a scalar, $\sigma$ is sigmoid function, and $W_g$ and $b_g$ are learnable parameters. The hyperparameter $\epsilon$ prevents the outliers
to break the memory values. 
The memory update at time step $t$ is formalized as:
\begin{equation}
    e_{k_{\rm best}}^{t+1} = g^t \cdot x^t + \left(1 - g^t\right) \cdot e_{k_{\rm best}}^t.
\end{equation}

\paragraph{Reading mechanism.} The reading mechanism characterizes
how the controller uses its memory and the current input to produce an
output. Instead of reading one memory cell, we aim to learn the 
support of all meta-evidence nodes. Thus, the
weighted sum is used to create a support vector
$s^{t}$:

\begin{equation}
    s^{t} = \sum_{k=1}^K \left(w_{e_k^{t}} \cdot e_k^{t+1}\right)
\end{equation}

We then incorporate the original input with the support vector $s^{t}$ to produce the negative/positive class probabilities $P(x^{t})$ as follows:
\begin{equation}
    P(x^{t}) = {\rm softmax}\left(W_p \left(s^{t} + x^{t}\right) + b_p\right).
\end{equation}

\begin{figure*}[t]
    \centering
    \includegraphics[width=0.9\textwidth]{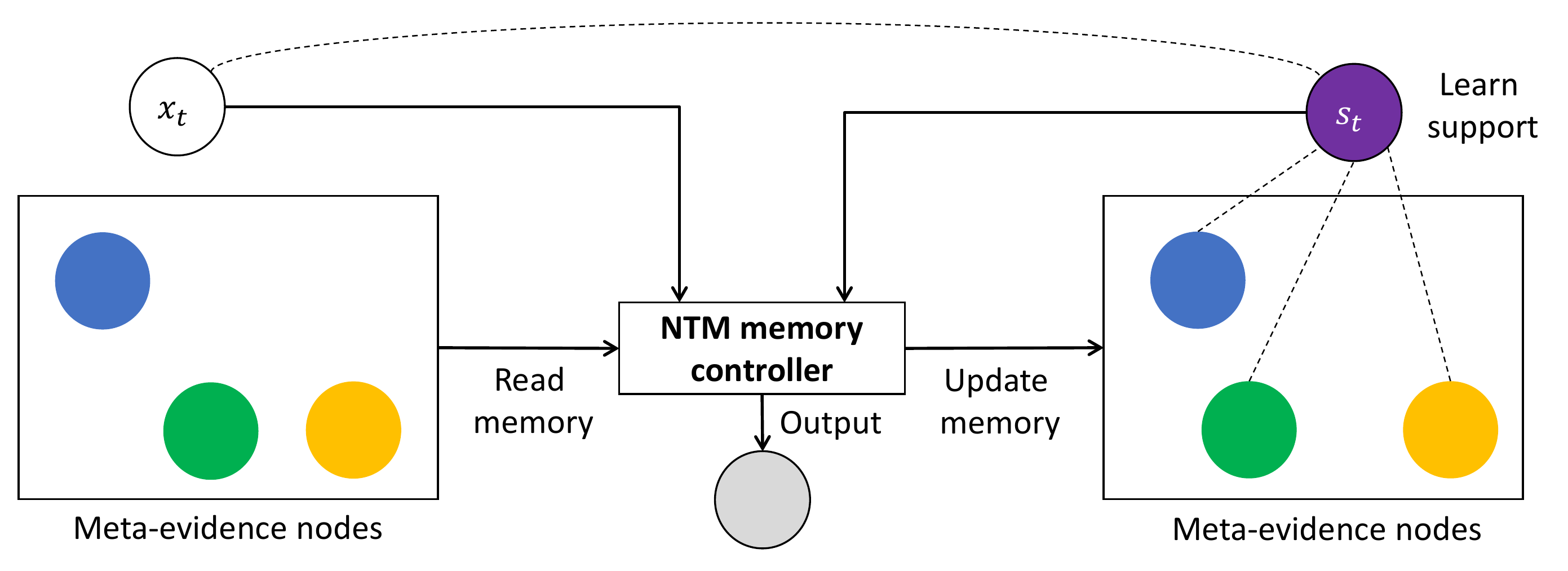}
    \caption{The information flow of our model.}
    \vspace{-5mm}
    \label{fig:architecture}
\end{figure*}

The overall information flow of our model is visualized in Figure~\ref{fig:architecture}. Our formulation draws inspiration from the NTM and the 
memory network. Our indexing algorithms in writing and reading mechanisms
are similar to the memory network, which is simpler than the NTM.
However, the memory network only stores 
intermediate computation steps in the memory, and these memories
can be considered as internal layers of the network. Our memory, on the contrary,
is external and not trained, only updated by the writing mechanism. In this regard, the memory bank of our model is more similar to the NTM.

\section{Experimental Results}
\label{sec:exp}
\subsection{Question answering performance}

In this subsection, we present our core results on two most popular datasets for answer selection: WikiQA \cite{Yang2015WikiQAAC} and TrecQA  \cite{wang2007jeopardy}. Due to space constraint, details of these datasets are described in the Appendix. Similar to previous work, we use two standard measures for the task: mean average precision (MAP) and mean reciprocal rank (MRR). Our models make use of the RoBERTa contextual embedding \cite{Liu2019RoBERTaAR}, pretrained on the ANSQ dataset~\cite{garg2019tanda}. For our model, we vary the number of memory cells from 2 to 64. The base configuration with 2 memory cells mimics the prototypical 
network with one cell for each prototype class representation.
Table~\ref{tab:TrecQA_WikiQA_overall} shows our model and the
baselines' performance. All our model's configurations outperform
the previous state-of-the-art models.\footnote{Due to space constraint, more details about the baselines can be found in the Appendix.} Increasing the number of memory cells
beyond the basic 2 cells - one for each class - clearly helps. The performance peaks at 16 or 32 cells depending on the dataset.

\newcommand{\modelacronym}{Evidence memory}
\begin{table*}[!t]
\small{
\centering
{
\begin{tabular}{L{0.4\columnwidth}|C{0.1\columnwidth}C{0.1\columnwidth}|C{0.1\columnwidth}C{0.1\columnwidth}}
\hline
\multirow{3}{*}{\textbf{Model}}   & \multicolumn{2}{c|}{\textbf{TrecQA}}  & \multicolumn{2}{c}{\textbf{WikiQA}} \\ 
\cline{2-5}              & \multicolumn{1}{c}{MAP}   & \multicolumn{1}{c|}{MRR}   & \multicolumn{1}{c}{MAP}   & \multicolumn{1}{c}{MRR} \\
\hline 
RoBERTa + Evidence Memory (2) (ours) & {0.949} & {0.982} & 0.925 &  0.940 \\



RoBERTa + Evidence Memory (16) (ours) & {0.954} & {0.982} & \textbf{0.936} & \textbf{0.952} \\

RoBERTa + Evidence Memory (32) (ours) & \textbf{0.961} & \textbf{0.993} & {0.933} & {0.945} \\

RoBERTa + Evidence Memory (64) (ours) & {0.949} & {0.974} & 0.929 & 0.941 \\

\hline
RoBERTa + CETE \shortcite{laskar2020contextualized} & {0.936} & {0.978} & {0.900} & {0.915} \\
RoBERTa + ANSQ Transfer \shortcite{garg2019tanda} & {0.943} & {0.974} & {0.920} & {0.933} \\
BERT + GSAMN + Transfer \shortcite{lai2019gated} & {0.914} & {0.957} & {0.857} & {0.872}   \\
IWAN + sCARNN \shortcite{Tran2018TheCA} & 0.829 & 0.875 & 0.716 & 0.722 \\
Compare-Aggregate \shortcite{Bian2017ACM} & 0.821 & 0.899 & 0.748 & 0.758 \\
\hline
\end{tabular}
}
}
\caption{Question answering performance.}
\vspace{-3mm}
\label{tab:TrecQA_WikiQA_overall}
\end{table*}

\subsection{Error-tracing performance}

One of the main motivations behind our evidence-based model is the 
ability to interpret the output of the neural network. It is hard
to quantify the interpretability of different models, however. To 
create a benchmark for interpretability, we look for a potential
application of interpretability in real-life development of a deep neural network.

Data collection is one of the most important parts of a machine 
learning model's development cycle. In many cases, nevertheless, the
collected data is not always clean and consistent, either due to
errors made by annotators or equivocal data points. For example,
the popular Switchboard Dialog Act dataset~\cite{stolcke2000dialogue}
only has 84\% inter-annotator agreement. Thus, we would like to 
test how well different models help in identifying noisy instances
in the dataset.

Our model naturally learns the most supportive group of instances
given a new instance, and thus, we can easily use this information
to trace from an error in validation to a group of training 
instances. Ideally, we will need to test all the training samples
of that group, but that would quickly make the number of
samples we need to check get out of control. Hence, we rely on  
heuristics: from the most relevant group, we only test the top $k$ most similar
instances (by cosine distance in the embedding space).
To create a noisy dataset given our current QA datasets, we randomly
swap 10\% the labels in each training set.\footnote{We tested with
other percentages, but we deem 10\% to provide the best 
compromise. Lowering the error rate, the model makes very few 
mistakes,
thus the statistics are unreliable. Increasing the error rate, the 
model performs too badly, then the tracing performance would be almost
perfect, but not meaningful.} We then calculate the percentage
of errors in validation that the model can 
correctly trace back to the training set perturbation. For quantitative
benchmark,
we compare our proposed model with the best baseline (i.e. the RoBERTa 
+ ANSQ transfer model) and the top $k$ 
most similar representations.

\begin{table*}[!ht]
\small{
\centering
{
\begin{tabular}{L{0.4\columnwidth}|C{0.1\columnwidth}C{0.1\columnwidth}|C{0.1\columnwidth}C{0.1\columnwidth}}
\hline
\multirow{3}{*}{\textbf{Model}}   & \multicolumn{2}{c|}{\textbf{TrecQA}}  & \multicolumn{2}{c}{\textbf{WikiQA}} \\ 
\cline{2-5}              & \multicolumn{1}{c}{Top 1}   & \multicolumn{1}{c|}{Top 3}   & \multicolumn{1}{c}{Top 1}   & \multicolumn{1}{c}{Top 3} \\
\hline 

Random & 0.10 & 0.27 & 0.10 & 0.27 \\
RoBERTa  + ANSQ Transfer~\shortcite{garg2019tanda} & {0.19} & {0.50} & {0.20} & {0.59} \\
RoBERTa + Evidence Memory (16) (ours) & {0.53} &  {0.81} & \textbf{0.51} & {0.90} \\
RoBERTa + Evidence Memory (32) (ours) & \textbf{0.80} & \textbf{0.88} & \textbf{0.51} & \textbf{0.93} \\
\hline
\end{tabular}
}
\caption{Error-tracing precision.}
\label{tab:trace}
}
\vspace{-3mm}
\end{table*}

Table ~\ref{tab:trace} shows the error-tracing performance of the 
model compared to the baseline. Our best model shows strong
error-tracing capability and outperforms the baseline by a wide margin.
On both datasets, our model can trace roughly 90\% of the errors
to the perturbed data points. This experiment clearly shows that
forcing a model to provide direct evidences helps in identifying
noisy training instances.


\section{Conclusion}
\label{sec:conclusion}
In this paper, we propose a novel neural network architecture that not
only achieves state-of-the-art performance on popular QA datasets,
but also shows strong error-tracing performance, which we argue will
be of great benefits to real-life machine learning applications. In the
future, we would like  to apply the model on different noisy user-generated datasets to test and further improve its interpretability.


\bibliographystyle{coling}
\bibliography{coling2020}

\newpage
\appendix
\section{Evaluated Datasets Description}
In this work, we evaluate the effectiveness of our proposed model on two datasets: TrecQA and WikiQA. The TrecQA datasets~\cite{wang2007jeopardy} is one of the most popular benchmarks for Answer Selection. The questions and answer candidates from this 
dataset is collected from the Text REtrieval Conference (TREC) 8–13 QA
dataset. In the original work, Wang et al.~\shortcite{wang2007jeopardy} 
used the questions from TREC 8-12 for training and the question from 
TREC 13 for test and validation set. In recent literature, most works
adopted the clean version~\cite{wang2015faq,tan2015lstm} of the
dataset, in which question with no answers or only with 
positive/negative answers are removed from the validation and
the test set. This version has 1,229/65/68 questions and 
53,417/1,117/1,442 question answer pairs for the train/dev/test sets. The WikiQA dataset \cite{Yang2015WikiQAAC} is constructed from user queries in Bing and Wikipedia. Similar to previous works
\cite{Yang2015WikiQAAC,Bian2017ACM,Shen2017InterWeightedAN}, we 
removed all questions with no correct answers before training and 
evaluating answer selection models. The excluded WikiQA has 
873/126/243 questions and 8,627/1,130/2,351 question-answer pairs for 
train/dev/test split.

\section{Answer Selection Baselines}
In this work, we compare the performance of our proposed models with several state-of-the-art models for answer selection, including \textit{Compare-Aggregate} \cite{Bian2017ACM}, \textit{IWAN + sCARNN} \cite{Tran2018TheCA}, \textit{BERT + GSAMN + Transfer} \cite{lai2019gated}, \textit{RoBerTa + ANSQ Transfer} \cite{garg2019tanda}, and \textit{RoBerTa + CETE} \cite{laskar2020contextualized}. The baselines \textit{Compare-Aggregate} and \textit{IWAN + sCARNN} employ the Compare-Aggregate architecture which had been used extensively before the appearance of large-scale pretrained language models such as ELMo \cite{Peters2018DeepCW} or BERT \cite{devlin2018bert}. Under the Compare-Aggregate architecture, small units of the input sentences are first compared and aligned. These comparison results are then aggregated to compute a final score indicating how relevant the candidate is to the question. More recent models, \textit{BERT + GSAMN + Transfer} \cite{lai2019gated}, \textit{RoBerTa + ANSQ Transfer} \cite{garg2019tanda}, and \textit{RoBerTa + CETE} \cite{laskar2020contextualized} are built upon the popular large-scale pretrained 
contextual embedding such as RoBerTa and Bert, and employ transfer 
learning from external larger corpus to achieve higher performance.

\end{document}